\documentclass[conference,10pt]{IEEEtran}
\usepackage[letterpaper, left=0.625in, right=0.625in, top=0.75in, bottom=1.0in]{geometry}

\usepackage{booktabs}
\usepackage{stix}

\usepackage{amsfonts}
\usepackage{xcolor,soul,framed} %,caption

\colorlet{shadecolor}{yellow}
\usepackage[pdftex]{graphicx}
\graphicspath{{../pdf/}{../jpeg/}}
\DeclareGraphicsExtensions{.pdf,.jpeg,.png}

\usepackage[cmex10]{amsmath}
\usepackage[T1]{fontenc}

\usepackage{array}
\usepackage{mdwmath}
\usepackage{mdwtab}
\usepackage{eqparbox}
\usepackage{url}
\usepackage{blkarray}

\usepackage{amsmath, amsfonts, amssymb}
\usepackage{mathtools}

\usepackage{algorithm}
\usepackage{algpseudocode}
\usepackage{float} 
\usepackage{subcaption}
\usepackage[T1]{fontenc}  % 字体编码要加 T1

\usepackage{etoolbox}
\makeatletter
\patchcmd{\ALG@doentity}{\noindent}{}{}{}
\makeatother
\usepackage{mathrsfs}

\usepackage[colorlinks=true, linkcolor=black, citecolor=black, urlcolor=black, bookmarks=false]{hyperref}

\algnewcommand\Output{\textbf{Output: }}
\begin{document}
 \setlength{\columnsep}{0.21in}
 \bstctlcite{IEEEexample:BSTcontrol}
     \title{Cluster-Based Client Selection for Dependent Multi-Task Federated Learning in Edge Computing}
%  

% ====================================================================

\author{\IEEEauthorblockN{Jieping Luo\IEEEauthorrefmark{1}, Qiyue Li\IEEEauthorrefmark{2}, Zhizhang Liu\IEEEauthorrefmark{2},  Hang Qi\IEEEauthorrefmark{2}, Jiaying Yin\IEEEauthorrefmark{3}, Jingjin Wu\IEEEauthorrefmark{2}\\}

\IEEEauthorblockA{\IEEEauthorrefmark{1}Department of Statistics, University of Oxford, 24-29 St Giles', Oxford OX1 3LB, U.K. \\}
\IEEEauthorblockA{\IEEEauthorrefmark{2}Guangdong Prov./Zhuhai Key Lab of IRADS, Beijing Normal-Hong Kong Baptist University, Guangdong, P. R. China \\}
\IEEEauthorblockA{\IEEEauthorrefmark{3}Institute of Precision Medicine, The First Affiliated  Hospital, Sun Yat-Sen University, P. R. China\\}
Email: jieping.luo@reuben.ox.ac.uk; t330005026@mail.uic.edu.cn;
t330031181@mail.uic.edu.cn;
\\
t330202706@mail.uic.edu.cn; yinjy35@mail.sysu.edu.cn;
jj.wu@ieee.org\\}

\maketitle

% === ABSTRACT ====================================================================
% 
%%%=================================================================================
\begin{abstract}
We study the client selection problem in Federated Learning (FL) within mobile edge computing (MEC) environments, particularly under the dependent multi-task settings, to reduce the total time required to complete various learning tasks. We propose CoDa-FL, a Cluster-oriented and Dependency-aware framework designed to reduce the total required time via cluster-based client selection and dependent task assignment. Our approach considers Earth Mover's Distance (EMD) for client clustering based on their local data distributions to lower computational cost and improve communication efficiency. We derive a direct and explicit relationship between intra-cluster EMD and the number of training rounds required for convergence, thereby simplifying the otherwise complex process of obtaining the optimal solution. Additionally, we incorporate a directed acyclic graph-based task scheduling mechanism to effectively manage task dependencies. Through numerical experiments, we validate that our proposed CoDa-FL outperforms existing benchmarks by achieving faster convergence, lower communication and computational costs, and higher learning accuracy under heterogeneous MEC settings.

\end{abstract}

% === KEYWORDS ====================================================================
% =================================================================================

% \fi
\IEEEpeerreviewmaketitle

% ====================================================================
% ====================================================================
% ====================================================================

% === I. INTRODUCTION =============================================================
% =================================================================================
\section{Introduction}

% MEC has been considered as a promising technology to enable mobile cloud computing, network control and storage~\cite{mao2017survey}. 
%%%
%%第一段不变【引用我们的ref】；
% 第二段稍微改下；
% 第三段小改；
% 第四段讲FL的过程，小改；

% qi2025energy

Federated learning (FL)~\cite{yang2019federated} is an emerging distributed machine learning paradigm that allows multiple clients to collaboratively train models while preserving data privacy. This is achieved by keeping the data local on devices and only sharing model updates. One notable application of FL is in mobile edge computing (MEC)~\cite{mao2017survey}, where an edge server provides data processing services for computation-intensive and latency-critical applications of edge clients, such as %augmented reality~\cite{siriwardhana2021survey}, unmanned aerial vehicles~\cite{sun2021joint}, 
satellite communications~\cite{qi2025energy}, supply chain demand forecasting~\cite{qi2025comparative}, and healthcare~\cite{datta2024blockchain}. One notable problem in MEC is network congestion caused by largely increasing numbers of devices, leading to various concerns including scalability and privacy. FL addressed these challenges by enabling most data processing and training tasks within mobile devices (FL clients), thereby avoiding transmitting raw data to MEC servers~\cite{bonawitz2019towards}. %Furthermore, the nature of MEC allows dynamical allocation of training tasks according to device capabilities for FL~\cite{10195234}, significantly reducing communication overhead and enhancing the efficiency of edge application execution. 

%While the FL process in MEC is similar to other general multi-task FL training scenarios, mobile edge clients face unique challenges in client heterogeneity problems and processing the inherent dependencies among different tasks, including the task itself and the interrelationships issues between executor deployment and task assignment to enhance the energy efficiency. Several approaches have been proposed to make better client selection to optimize the overall utility using different criteria such as system resources, network conditions, and data quality. One noteworthy problem for FL in MEC is to select suitable participating clients especially with multiple tasks, heterogeneous data distributions  and resource constraints

While the FL process in MEC shares many similarities with other general FL scenarios, mobile edge clients face unique challenges. Clients differ markedly in communication throughput and computational capacity; their local datasets are typically non-IID; and the scheduler must handle dependencies not only within each task but also across the coupled decisions of where tasks are placed and how they are allocated. Therefore, a key problem in MEC-based FL is the selection of appropriate clients, especially when dealing with multiple tasks, heterogeneous data, and limited resources~\cite{8761315}. A key challenge in addressing this problem is that, with many edge devices, exhaustive task–client assignment quickly becomes computationally prohibitive.

%Specifically, processing large datasets improves model accuracy but incurs substantial computational energy costs~\cite{9498853}, while clients with poor channel conditions require higher transmission power for reliable model updates particularly during packet retransmissions~\cite{ling2024efficient}. However, clients typically possess heterogeneous data across multiple tasks—such as physiological signals, motion data, or behavioral logs—with varying quantity and quality~\cite{10713971}. 

In this work, we propose CoDa-FL (Cluster-oriented and Dependency-aware Client Selection for Federated Learning), a framework designed to improve the overall efficiency of FL in MEC environments via client selection. Two key features that distinguish our proposed framework with existing similar studies are, 1) we explicitly account for the inherent dependencies between FL tasks, which is an important feature in MEC~\cite{10713971}; and 2) we consider cluster-based selection based on the data similarity among edge devices, in order to balance computational efficiency, communication reliability, and learning accuracy of FL.

As mentioned earlier, most prior work on FL client selection has focused on single-task settings. For example, FedCS~\cite{8761315} was introduced as a resource-aware protocol that maximizes the number of participating clients within a given training deadline. On the other hand, recent research on multi-task FL has largely assumed that tasks are independent. Zhou \textit{et al.}~\cite{zhou2022efficient}, for instance, proposed a framework for parallel training of multiple FL jobs using scheduling policies that assign devices to tasks to minimize overall cost. Similarly, cluster-based client selection policies have been explored in conjunction with task assignment, but again under the independence assumption~\cite{cheng2021joint}.

However, existing studies have not addressed client selection strategies that jointly consider task dependencies and data-driven clustering, leaving a gap in supporting dependent multi-task FL in MEC environments. Our proposed CoDa-FL aims to fill this gap by starting with a clustering mechanism based on the Earth Mover's Distance (EMD)~\cite{rubner2000earth}, measured by the similarity of their task-specific data distributions. This clustering mechanism not only reduces the complexity of the client selection problem, but also leads to more structured strategies. Then, we perform convergence analysis to theoretically derive the relationship between intra-cluster EMD and an upper bound on the number of training rounds required for each cluster-task pair. Finally, we incorporate a DAG-based task scheduling mechanism, ensuring that tasks are executed in a dependency-aware and communication-efficient manner throughout the FL process. 
% In this paper, we tackle a novel setting where multiple FL tasks have dependency relationships (modeled by a DAG) and must be jointly scheduled.

Our main contributions are summarized as follows.
\begin{itemize}
    \item  We propose the CoDa-FL framework to enhance the overall efficiency of FL in MEC. This framework considers both inherent dependency between FL tasks and cluster-based client selection among edge devices with data heterogeneity, effectively addressing a critical issue in task assignment in MEC-based FL with limited resources and dependency within and between tasks.
    
    \item We first perform EMD-based clustering to group clients with similar data distributions, to reduce the communication and computational complexity in client selection problems of FL in MEC due to the potentially large number of clients. Then, we identify and derive a key relationship that the upper bound of training rounds required for each cluster-task pair monotonically increases with the intra-cluster EMD, by conducting the convergence analysis and relaxing the original objective function. %Based on the finding that the upper bound increases with intra-cluster EMD, we then obtain a metric to incorporate the 
    %DAG-based cluster-task scheduling.
    \item We validate through numerical experiments that our proposed CoDa-FL outperforms other benchmarks with traditional client selection strategies in the total time to complete all learning tasks with different targets. Specifically, CoDa-FL reduces the total time by 5.05\% compared to the closest benchmark. CoDa-FL's ability to optimize the training efficiency and convergence speed across different tasks supports the robustness of EMD-based clustering and dependent multi-task scheduling strategy, making it an effective solution for task-to-client matching
    of FL in MEC with large number of edge devices.

\end{itemize}

The rest of this paper is organized as follows. Section II describes the key concepts and definitions in the system model. Section III analyzes the convergence performance to simplify the optimization problem. Section IV explains proposed algorithms in detail. Section V presents the numerical results. Finally, Section VI concludes the paper.

% Traditionally, the objective of client selection is to select the maximum number of clients who can derive and upload their local models before the deadline in each global i(teration. However, selecting more clients increases the energy consumption of the clients.
%三个文章，一句话概括三个；然后说对比，引出我们的xxx。都没有考虑xx问题，但是这个问题很重要，所以我们。。。

\begin{figure}[htp]
    \centering
    \includegraphics[width=0.75\linewidth]{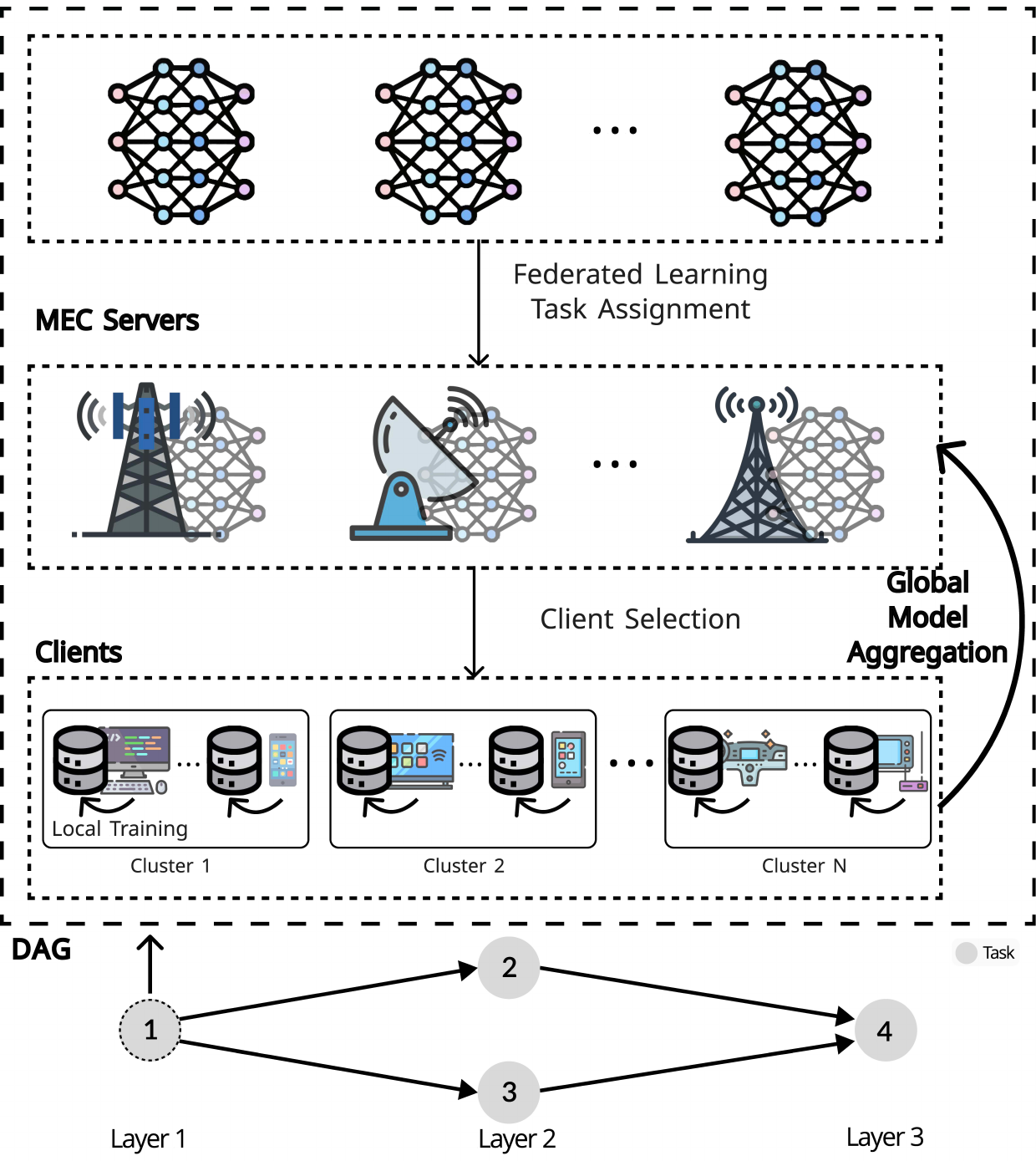}
    \caption{Cluster-based FL client selection in MEC.}    
    \label{fig:1}
\end{figure}
% Different federated learning tasks have various task requirements, and different clients have different data sets and are located in different environments. Thus, how to select suitable clients for different federated learning tasks and ensure the training performance in MEC networks presents a noteworthy problem.

% %关于训练，最开始的是从MEC server那里获取initial model
% Once the clients are selected by the MEC server, each client will receive an initialized model from the MEC server and
% perform local training to update the model parameters.

\section{System Model}
% 描述场景Edge computing，对xxxapplication，有用，为了dependency FL task ，用cluster来降低复杂度；有了cluster之后做convergence analysis； 方法是 结合EMD和DAG ； 发现越non-iid，收敛越慢。

\subsection{Dependent
Multi-Task Federated Learning}
We consider a federated learning (FL) system composed of a central (edge computing) server and $U$ client devices (mobile devices). Each client $u \in \mathcal{U}=\{1,2,...,U\}$ possesses a local dataset $D_{u}$ with size $|D_{u}|$ and participates in training via the classical FedAvg algorithm~\cite{li2019convergence}. The total data size across all clients is $|D| = \sum_{u \in \mathcal{U}} D_u$. We consider a collection of learning tasks organized as a directed acyclic graph (DAG), denoted as $G=(\mathcal{V},E)$, where each node $v\in \mathcal{V}$ represents an individual training task, and $\mathcal{V}=\{1,...,V\}$. Each task $v$ has a target accuracy $\tau_v$, and an initial global parameter $w_v^{0}$ stored at the respective edge server.

A directed edge $(v\to q)\in E$, where $v,q\in \mathcal{V}$, indicates that task $v$ must be completed (reaching the accuracy requirement) before task $q$ commences. The DAG is partitioned into $L$ layers, where each layer $l\in \mathcal{L}=\{1,2,...,L\}$ contains all tasks that can be executed in parallel. The execution proceeds sequentially from layer 1 to layer $L$. We define the set of tasks for each layer as $\mathcal{S}_{l} \in \mathcal{S}=\{\mathcal{S}_{1},...,\mathcal{S}_{L}\}$.

All clients in $\mathcal{U}$ are aggregated to $N$ disjoint clusters $\mathcal{C}=\{\mathcal{C}_1, \mathcal{C}_2,...,\mathcal{C}_N\}$, where each $\mathcal{C}_i \subseteq \mathcal{U}$, and $\mathcal{C}_i \cap \mathcal{C}_j = \emptyset$ for $i \neq j$. The number of clients in cluster $\mathcal{C}_i$ is $|\mathcal{C}_i|$. The clusters are formed based on data heterogeneity to be described in the next subsection. Each cluster $\mathcal{C}_i \in \mathcal{C}$ would be assigned for at most one task $v$ at the same time. 
We define a mapping $\phi: \mathcal{V} \to \{1, 2, \dots, N\}$ such that $\phi(v)$ indicates the index of the cluster assigned to task $v$.

In training round $r\in \{0,1,2,...,R_{v}\}$ where $R_{v}$ is the required communication rounds for task $v$ to meet $\tau_v$, the edge server holds the current global model parameters $w_v^{r}$. Each client in the selected cluster $
\mathcal{C}_{\phi{(v)}}$ for task $v$ downloads the global model $w_v^{r}$ and updates its local model. The global loss function $F_v(w)$ for task $v$ is a weighted sum of local losses across all clients in the assigned cluster $\mathcal{C}_{\phi(v)}$, given by
\begin{align}
    F_v(w) = \sum_{u \in \mathcal{C}_{\phi(v)}} 
    p_{v,u}  F_{v,u}(w),
\label{eq:global_obj}
\end{align}
where $p_{v,u}=\frac{|D_{u}|}{\sum_{k \in \mathcal{C}_{\phi(v)}} |D_k|}$ is the normalized data weight coefficient, and $F_{v,u}(w)$ is the local loss of model $w$ on client $u$ with respect to task $v$. We denote the global minimizer as
\begin{equation}
    w^*_v=\arg\min_w F_v(w).
\end{equation}
The local minimizer for client $u$ is
\begin{equation}
    w^*_{v,u}=\arg\min_w F_{v,u}(w).
\end{equation}
% $\mathcal{C}_{v,r}\subseteq \mathcal{C}$
In each round $r$, a cluster of clients is selected for task $v$, and all clients in the selected cluster perform $E$ steps of local SGD on $F_{v,u}$. For each task $v$, we define the corresponding accuracy as 
\begin{align}
    \epsilon_v = 1 - \frac{F_v(w_v^{R_v})}{F_v(w_v^{0})},
\end{align}
and we consider that the accuracy condition has been satisfied when
$\epsilon_v \ge \tau_v$.

% In convergence analysis we equivalently bound $F_i(w) - F_i(w^*)\le \varepsilon_i$.

\subsection{Data Heterogeneity}
%先介绍cluster，然后后面的时间Total time= cluster里面最慢的那个client（max ci）； 对应的client 和 cluster的EMD 怎么算;讲清楚time是怎么算的，删除所有与score有关的内容，只分析收敛和EMD 的关系
%定义cluster，每个cluster里面有多少个（不用强调是不是平均分）
In realistic MEC FL scenarios, the data distribution across clients is largely heterogeneous. %The evaluation of the distribution skewness is the key step to explore the impact of non-IID data on training for clients and FL server. 
The Earth Mover’s Distance (EMD) is commonly used as a metric to quantify data heterogeneity between each local dataset and the global dataset. Consider a supervised learning task $v$ with input features $\mathbf{x}$ and labels $y\in \mathcal{Y}=\{1,...,Y\}$, for each client $u$, the local data distribution follows $(\mathbf{x},y) \sim P_{u}(\mathbf{x},y)$. 
%We assume all local samples for each task $v$ follow the same global distribution and are independent of all other tasks, which can be represented as $(\mathbf{x},y) \sim P_{v}(\mathbf{x},y)$. 
Given the local label distribution $P_{u
}(y)$ and the global label distribution $P_{v}(y)$~\cite{hu2025faster}, the EMD of client $u$ for task $v$ is
\begin{align}
 \Delta_{u}= \sum_{y=1}^{Y} \left| P_{u}(y) - P_v(y) \right|.
\end{align}

% We quantify data heterogeneity using the Earth Mover’s Distance (EMD) between the client’s local data distribution $P_{i}$and the global data distribution~\cite{zhao2018federated}.
We compute EMD at the cluster level to reduce the computational cost and reflect the aggregated distribution of similar clients. Let %the label distribution of each cluster $\mathcal{C}_i$ be the weighted average of all the clients' distributions in the cluster, given by 
\begin{equation}
P_i(y) = \frac{1}{\sum_{u \in \mathcal{C}_i} |D_u|} \sum_{u \in \mathcal{C}_i} |D_u| \, P_u(y).
\end{equation}
The cluster-level EMD of cluster is
$\mathcal{C}_i$ is
\begin{equation}
\Delta_{\mathcal{C}_i} =\sum_{y=1}^{Y} \left| P_i(y) - P_g(y) \right|.
\end{equation}
Then, the average EMD across all $N$ clusters is 

\begin{equation}
    \begin{aligned}
\bar{\Delta}_{\mathcal{C}} &=\sum_{i=1}^{N} \frac{\sum_{u \in \mathcal{C}_{i}} |D_u|}{|D|} \Delta_{\mathcal{C}_{i}}.
% \\ & =\sum_{i=1}^{N} \frac{\sum_{u \in \mathcal{C}_{i}} |D_{c}|}{|D|} \sum_{y=1}^{Y}\left|P_{i}(y)-P_{g}(y)\right|,
\end{aligned}
\end{equation}
We define the round-level average heterogeneity $\bar{\Delta}_{v}^{r}$ for task $v$ in round $r$ as
\begin{equation}
\
\bar{\Delta}_{v}^{r} = \sum_{} \frac{|D_{\mathcal{C}_i}|}{|D_{v}^{r}|} \Delta_{\mathcal{C}_i}, 
\quad \text{where } |D_{v}^{r}| = \sum_{\mathcal{C}_i \in \mathcal{C}_{\phi{(\mathcal{V})}}^{r}} |D_{\mathcal{C}_i}|.
\end{equation}
The global data distribution for task $v$ is
\begin{equation}
P_{g_{(v)}}=\sum_{u\in \mathcal{C}_{\phi(v)}} p_{v,u}P_{v,u}.
\end{equation}

% Let $P_{g_{(v)}}=\sum_{u\in \mathcal{C}_{\phi(v)}} p_{v,u}P_{v,u}$ be the global data distribution for task $v$. 
We consider $\Delta_u = W_1(P_{v,u},P_{g_{(v)}})$ for each client $u$ as a measure of how far client $c$’s local distribution is from the global distribution. And we denote $q_{v,\mathcal{C}_i}=\sum_{u\in \mathcal{C}_i}p_{v,u}$ as the weighted coefficient of cluster $i$ for task $v$. Then the overall heterogeneity across all clusters is 
\begin{equation}
\Delta_{\mathcal{C}} = \sum_{i=1}^{N} q_{\mathcal{C}_{i,v}}\Delta_{\mathcal{C}_i}.
\end{equation}
% \vspace*{0.7cm} 
% where $q_{v,\mathcal{C}_i}=\sum_{u\in \mathcal{C}_i}p_{v,u}$ is a weighting coefficient of cluster $i$ for task $v$.

% For each cluster $\mathcal{C}_i\in \mathcal{C}$, let $P_{\mathcal{C}_{i,v}}$ be the aggregated distribution of cluster $\mathcal{C}_i$. 

% \subsection{Dependency of multiple tasks}

% %In our system, we divide the total tasks into a sequence of tasks organized in a directed acyclic graph (DAG) structure. These tasks should be processed strictly in the specific logical precedence sequence. Each task $v$ corresponds to training a model to a required accuracy level $\tau_v$ before its dependent tasks can proceed. 

% We use a directed acyclic graph (DAG), $G=(\mathcal{V},E)$, to represent the dependency of tasks. Each node $v\in \mathcal{V}$ in the DAG represents a task and a directed edge $(v\to q)\in E$, where $v,q\in \mathcal{V}$, indicates that task $v$ must be completed (reaching the accuracy requirement) before task $q$ commences. The DAG is partitioned into $L$ layers, where each layer $l\in \mathcal{L}=\{1,2,...,L\}$ contains all tasks that can be executed in parallel. The execution proceeds sequentially from layer 1 to Layer $L$. We define the set of tasks for each layer as $\mathcal{S}_{l} \in \mathcal{S}=\{\mathcal{S}_{1},...,\mathcal{S}_{L}\}$. Then 

\subsection{Transmission and Computation Time}

In each FL training round, the latency (total time) incurred by client $u$ is composed of the local computation time required to process local updates and the uplink communication time to upload the model to the server~\cite{liu2023dependent}.

% Given a computational complexity $\kappa$ and a device-specific computing capacity $f_u$, the local computation time at client $u$ for round $r$ is
% \begin{equation}
% t^{(r),\text{comp}}_u = \frac{E|D_u|\kappa}{f_u}.
% \label{eq:comp_time}
% \end{equation}
%where $f_u=\frac{1}{C_{|D_u|}} f^{\text{clock}}_u$, and $C_{|D_u|}$ is the number of CPU cycles per bit required to compute 1 bit of data, and $f^{\text{clock}}_u$ is the CPU clock frequency of client $u$.
% quantifies the ratio between the power of a desired signal and the power of background noise.
% The Signal-to-Noise Ratio (SNR) 
% \begin{equation}
%     \text{SNR} = \frac{p_c \cdot h_{c}}{\sigma^2},
% \end{equation}
% where $p_c$ is the transmission power of client $c$, $h_c$ is the channel gain from client $c$ to the server, $\sigma^2$ is the background noise power (in watts). 
%For each client $u$, the Signal-to-Noise Ratio (SNR) for uplink transmission is $\text{SNR} = \frac{p_u \cdot h_{u}}{\sigma^2}$, where $p_u$ is the transmission power, $h_u$ is the uplink channel gain, $\sigma^2$ is the thermal noise. 
By Shannon's capacity formula, the achievable transmission rate is
\begin{equation}
\begin{aligned}
R^r_u = B \log_2 \left(1 + \text{SNR} \right)=B \log_2 \left(1 + \frac{b^r_u  h_{u}}{\sigma^2} \right),
\label{eq:rate}
\end{aligned}
\end{equation}
where $B$ is the channel bandwidth, $b^r_u$ is the transmission power, $h^r_u$ is the uplink channel gain, $\sigma^2$ is the thermal noise. 
The total time $t^r_u$ at round $r$ for client $u$ is
\begin{equation}
t^{r}_u = \frac{E|D_u|\kappa}{f_u}+  \frac{S^r_u}{R^r_u}, 
% \begin{aligned}
% t_u &= t^{\text{comp}}_u + t^{\text{comm}}_u \\
% &= \frac{E|D_u|\kappa}{f_u} + \eta_u \frac{S}{R_u},\\
% &= \frac{E|D_u|\kappa}{f_u} + \eta_u  \frac{S}{B \log_2 \left(1 + \frac{p_u h_u}{\sigma^2} \right)}.
% \end{aligned}
\label{eq:client_latency}
\end{equation}
where $S^r_u$ denotes the model size for the task that $u$ is assigned to (in bits) in round $r$, $\kappa$ is the computational complexity coefficient, and $f_u$ is a device-specific computing capacity. 

The total time required to complete task $v$ is
\begin{equation}
    T_v=\sum_{r=1}^{R_v}T_{v}^{r} = \sum_{r=1}^{R_v}\max_{u \in \mathcal{C}_{\phi(v)}} t_u^{r}.\label{eq:total_time}
\end{equation}
% The total time for completing all tasks is,
% \begin{equation}
%   T_{\mathrm{total}}
%   \;=\;\sum_{l=1}^{L} T_{l}
%   \;=\;\sum_{l=1}^{L}\max_{\,v\in\mathcal{S}_{l}}\,T_{l,v},
% \end{equation}
% where $T_{l}$ is the transmission time for each layer $l$ in the DAG, $T_{l,v}$ is the time for each task $v$ in layer $l$.

\subsection{Optimization Problem}
% 111
We aim to minimize the total time to complete all learning tasks. %We partition the DAG into L layers, where each layer contains tasks that can be executed in parallel. The total training time is determined by the sequential execution of layers, and the time for each layer is dominated by the longest task within it. We use a cluster-level abstraction for scheduling, but the round latency is determined by the slowest participating client within selected clusters.
Formally, the objective is
\begin{equation}
    \begin{aligned}
\min_{{\mathcal{G}{i,r}}} \quad &  T_{\mathrm{total}}
  \;=\;\sum_{l=1}^{L} T_{l}
  \;=\;\sum_{l=1}^{L}\max_{\,v\in\mathcal{S}_{l}}\,T_{l,v}, \\
    &    \text{s.t.} \quad \epsilon_v \ge \tau_v, \quad \forall v \in \mathcal{V},
        \label{eq:cluster_obj}
\end{aligned}
\end{equation}
% sum每一层的时间，每一层的时间=max这层最长的task的时间，这层最长的task时间=sum该task下训练的rounds下的时间；每一轮的训练时间=max clients中最长的时间；
where $T_{l}$ is the transmission time for each layer $l$ in the DAG, and $T_{l,v}$ is the time for each task $v$ in layer $l$.

\section{Convergence Analysis}

To minimize the total time across all tasks with complex data heterogeneity among clients, intra-cluster variations and different task objectives, directly solving the optimization problem is challenging. Therefore, we conduct a convergence analysis for each task $v \in \mathcal{V}$ to simplify the problem.

% Recall that, as defined in~\eqref{eq:global_obj}, the cluster-level loss for task $v$ with parameters $w$ is $F_ v(w)$. 
We consider that the local loss functions for all tasks 1) have bounded the variance of stochastic gradients, and 2) have bounded distribution divergence.

Our goal is to find the minimum number of training rounds $R_v$, and the expected optimality gap for task $v$ satisfies
\begin{equation}
\mathbb{E}\left[F_v(w^{R_v}) - F_v(w_v^*)\right] \le \varepsilon_v.
\end{equation}
We denote $w^{r}$ as the global model after round $r$, and the global update for the next round is
\begin{equation}
w^{r+1} = \sum_{u \in \mathcal{C}_{v,r}} p_{v,u} w_u^{r+1}.
\end{equation}

By the bounded variance condition, the variance of stochastic gradients over $U$ clients satisfies, denoted as $\mathcal{I}$,
\begin{equation}
\mathcal{I} \le \frac{1}{U} \sum_{c \in \mathcal{C}_{v,r}} p_{v,u}^2 \sigma_{v,u}^2 \le \frac{\sigma^2}{U}.
\end{equation}
The client drift $\mathcal{D}$ from multiple local steps is bounded by
$\mathcal{D} \le (E-1)^2 G^2$. We assume $|\nabla F_{v,u}(w)| \le G$ for all $u$ and $w$, where $G > 0$ is a constant. The client drift term is shown to grow with $E$ and the gradient norm $G$~\cite{mcmahan2017communication}. We denote $\Gamma_v$ as the total optimality gap for task $v$, which is the weighted sum of these gaps over all clients or clusters, given by 
\begin{equation}
\begin{aligned}
\Gamma_v &= \sum_{u\in \mathcal{C}_{\phi(v)}} p_{u,v}\Big(F_v(w_v^*) - F_{v,u}(w_{v,u}^*)\Big) \\
&= \sum_{i=1}^{N} q_{v,\mathcal{C}_i}\Big(F_v(w_v^*) - F_{v,\mathcal{C}_i}(w_{\mathcal{C}_{v,\mathcal{C}_i}}^*)\Big).
\label{eq:Gamma_def}
\end{aligned}
\end{equation}

By the bounded distribution divergence condition, we have
\begin{equation}
    \Gamma_v \le \sum_{i}q_{v,\mathcal{C}_i}L_d\Delta_{\mathcal{C}_i} = L_d\sum_{i}q_{v,\mathcal{C}_i}\Delta_{\mathcal{C}_i} = L_d\Delta_{\mathcal{C}}^{v},
\end{equation}
where $\Delta_{\mathcal{C}}^{(v)}$ is the overall cluster-level EMD for task $v$. Intuitively, $\Gamma_v=0$ if all clients’ distributions are identical to the global distribution for IID data, and $\Gamma_v$ grows as data becomes more non-IID. Then the expected progress is 
\begin{align}
\mathbb{E}\big[F_v(w^{r+1}) - F_v(w_v^*)\big] 
&\le (1 - \mu\eta E)^r \big(F_v(w^{0}) - F_v(w_v^*)\big) \nonumber\\
&\quad + \Gamma_v + \mathcal{I} + \mathcal{D}.
\end{align}
Let $\rho = 1 - \mu \eta E$, then we can get
\begin{align}
\mathbb{E}\big[F_v(w^{R}) - F_v(w_v^*)\big] 
&\le \rho^R \big(F_v(w^{0}) - F_v(w_v^*)\big) + \Gamma_v\nonumber\\
&\quad + \frac{\sigma^2}{U} + (E - 1)^2 G^2 \le \varepsilon_v.
\end{align}
Solving for $R_v$,
\begin{equation}
R_v(\varepsilon_v) \ge \frac{1}{\mu \eta E} \cdot \log \left( \frac{\Delta_0}{\varepsilon_v - \left(\Gamma_v + \frac{\sigma^2}{U} + (E - 1)^2 G^2\right)} \right).
\end{equation}
Hence, we have
\begin{equation}
    R_v(\varepsilon_v) \propto \frac{1}{\varepsilon_v} \left( \Gamma_v + \mathcal{I} + \mathcal{D} \right).
\end{equation}
Using $\Gamma_v \le L_d \sum_i q_{v,i} \Delta_{\mathcal{C}_i}$ and rearranging, we obtain
% \begin{align}
% \begin{equation}
% R_v(\varepsilon_v) = \mathcal{O}\left( \frac{E G^2 + \sum_u p_{v,u}^2 \sigma_{v,u}^2 + L_d L \sum_i q_{v,\mathcal{C}_i} \Delta_{\mathcal{C}_i} + (E - 1)^2 G^2}{\mu \varepsilon_v} \right).
% \label{equ: final}
% \end{equation}
% \end{align}

\begin{equation}
\begin{split}
R_v(\varepsilon_v) &=  \mathcal{O}\left( \frac{E G^2 + \sum_u p_{v,u}^2 \sigma_{v,u}^2 + L_d L \sum_i q_{v,\mathcal{C}_i} \Delta_{\mathcal{C}_i}}{\mu \varepsilon_v} \right. \\
& \left. + \frac{(E - 1)^2 G^2}{\mu \varepsilon_v} \right).
\label{equ: final}
\end{split}
\end{equation}

Equation~\eqref{equ: final} indicates the relationship between the number of training rounds required to reach specific $\varepsilon_v$ for each task $v$, and cluster-level EMD clustering $\Delta_{\mathcal{C}_i}$ due to data heterogeneity among clients. The bound explicitly shows that $R_i(\varepsilon_v)$ grows linearly with the average EMD across clusters. In particular, the term $L_d L \sum_i q_{v,\mathcal{C}_i} \Delta_{\mathcal{C}_i}$ quantifies how statistical heterogeneity slows convergence. Therefore, selecting clusters with lower $\Delta_{\mathcal{C}_i}$ values reduces the convergence bound, thereby highlighting a cluster-aware client selection strategy that prioritizes distributional proximity to the global data.

% which shows a cluster-aware client selection strategy that favors distributional proximity to the global data.
% When all clients are IID (i.e., $\Delta_c = 0$ for all $c$), the heterogeneity term vanishes. The convergence rate then matches that of centralized SGD, up to variance and drift terms.
\section{Algorithms}
\subsection{EMD-Based Client Clustering}
Based on the observation that the upper bound increases with the average EMD across clusters, we identify intra-cluster EMD as a critical parameter in addressing the optimization problem of minimizing the total learning task time. Based on this insight, we propose an algorithm to efficiently solve the objective in equation~\eqref{eq:cluster_obj} by computing the cluster-level EMD and performing client clustering, as detailed in Algorithm~\ref{algorithm:1}.
\begin{algorithm}[H]
\caption{EMD-based Client Clustering}
\label{algorithm:1}
\begin{algorithmic}[1]
\Require Client label histograms $\{h_u\}_{u=1}^U$, number of clusters $N$, total number of clients $U$
% \Output Clusters $\mathcal{C} = \{N_1, N_2, \dots, N_C\}$

%\State \textbf{Normalize:} 
\For{$u = 1$ \textbf{to} $U$}
    \State Convert the client label histogram $h_u$ to $p_u = \frac{h_u}{\sum_{k=1}^d h_u[k]}$
    \EndFor

% \State \textbf{Compute Distance Matrix:} 
\For{$u, u' = 1$ \textbf{to} $U$}
    \State Compute the pairwise EMD $D_{u,u'} = \|p_u - p_{u'}\|_1$
    \EndFor

% \State \textbf{Cluster:} 
\State Assign cluster labels $\ell_u \in \{1, \dots, N\}$ to each client by applying agglomerative clustering to $D$

% Perform agglomerative clustering using the distance matrix $D$. The procedure assigns a label $\ell_u \in \{1, \dots, N\}$ to each client based on $D$.

% \State \textbf{Build Clusters:} 
\For{$i = 1$ \textbf{to} $N$}
    % \State Assign clients to the cluster based on their labels $\mathcal{C}_i \gets \{ u \mid \ell_u = i$\}
    \State Assign clients to clusters $\mathcal{C}_i \gets \{ u \mid \ell_u = i \}$
    \EndFor

\State \textbf{Output:} The final clusters $\mathcal{C}=\{\mathcal{C}_1,\mathcal{C}_2,\dots,\mathcal{C}_N\}$

\end{algorithmic}
\end{algorithm}
\noindent

This method groups clients with similar label distributions into the same cluster, enabling task-aligned model training and reducing negative transfer across unrelated tasks. Specifically, each client's local label histogram is first normalized into a probability distribution, ensuring fair comparison across clients with varying dataset sizes. The pairwise distances between all clients are then computed using the $\mathcal{L}_1$-norm, which corresponds to the EMD between categorical distributions. Subsequently, we apply agglomerative hierarchical clustering with average linkage on the precomputed distance matrix. Clients assigned to the same cluster are used to jointly train task-specific models. This clustering strategy serves as the foundation for our multi-layer scheduling framework. It ensures that each federated round involves homogeneous client groups, thereby accelerating convergence, reducing communication waste, and ultimately improving generalization across all downstream tasks.

While a non-cluster-based client selection policy can theoretically achieve shorter total completion time, its computational cost is prohibitive. Specifically, with $V$ tasks and $U$ clients, the exhaustive matching requires a search space size of approximately $\frac{V^U}{V!}$, which is practically impossible to enumerate. In contrast, our proposed EMD-based clustering only requires pairwise distance calculations between clients of complexity $O(U^2Q)$ (with $Q$ denoting the number of classes), followed by hierarchical clustering with $O(U^2\log U)$. After identifying clusters, the complexity of the scheduling problem is $O(NV)$, where $N\ll U$. %In summary, our proposed EMD-based clustering achieves much lower complexity without a severe sacrifice in performance, making it a more efficient and feasible solution in real-world FL scenarios.

\vspace{-1mm}

\subsection{Reinforcement Learning-Based DAG Scheduling}
To address the challenges of minimizing the overall training latency in a DAG-structured set of FL tasks, we consider the task dependency graph $\mathcal{G}$, the processing time matrix $\texttt{proc\_time}$ for each task-cluster pair as calculated by~\eqref{eq:rate} to~\eqref{eq:total_time}, and the execution status of both tasks and clusters. The state representation comprises the current state of all tasks, capturing whether they are not ready, ready, running, or completed, along with the real-time occupancy of each cluster. The detailed steps of the scheduling algorithm are demonstrated in Algorithm~\ref{algorithm:2}.

\begin{algorithm}[H]
\caption{RL-based DAG Task Scheduling with Cluster-level Latency Awareness}
\label{alg:RL-DAG}
\begin{algorithmic}[1]
\Require DAG structure $\mathcal{G}$, processing time matrix $\texttt{proc\_time}$, accuracy thresholds $\{\tau_i\}$
\State Initialize environment $\mathcal{E}$ with task and cluster states
\State Initialize PPO policy $\pi_\theta$ with neural network parameters
\For{each episode}
    \State Reset environment to initial state $s_0$
    \For{each time step $t$}
        \State Sample action $a_t \sim \pi_\theta(\cdot|s_t)$
        \State Execute $a_t$, observe $s_{t+1}$, $r_t$, and done flag
        \State Update PPO policy
        \If{all tasks completed}
            \State Record $T_{\text{total}}$ and task-cluster assignments
            \State Update the best solution if $T_{\text{total}}$ is improved
            \State \textbf{break}
        \EndIf
    \EndFor
\EndFor
\State \textbf{Output:} Optimal task-cluster assignment with min $T_{\text{total}}$
\end{algorithmic}
\label{algorithm:2}
\end{algorithm}
\noindent

 For training, we employ the Proximal Policy Optimization (PPO) algorithm, which iteratively refines the policy $\pi_\theta$ to maximize the expected cumulative reward. The PPO agent learns to dynamically allocate tasks to clusters in a manner that exploits parallelism while respecting task dependencies, thereby minimizing the critical path duration. To monitor training convergence and assess the quality of generated schedules, a custom callback tracks the makespan and the complete task-to-cluster assignment mappings. This ensures that the final scheduling policy simultaneously minimizes the total latency $T_{\text{total}}$ and meets the accuracy thresholds $\tau_i$ for each task. %The detailed pseudocode for the  RL-based DAG scheduling approach is provided in Algorithm~\ref{alg:RL-DAG}.

\section{Numerical Experiment}

\subsection{Experimental Setup}
%In our experiments, we simulate a set \(\mathcal{U}\) of clients whose system and channel characteristics are drawn from realistic distributions. 

We assume that each client \(u\in\mathcal{U}\) has a CPU frequency
$f_u \sim U[1.2\,\mathrm{GHz},\,2.5\,\mathrm{GHz}],$
and experiences a wireless channel power gain $h_u \sim \mathrm{Exp} (\lambda=2.5\times10^{-7})$. We fix the bandwidth at \(B=20\,\mathrm{MHz}\) and the model sizes for all tasks at \(S=0.5\,\mathrm{M}\)bits. The receiver noise power is set to \(\sigma^{2} = -43\,\mathrm{dBm}\), the transmit power per client is \(p_{u}=0.2\)W, and the computation cost per bit is \(C_{D_{u}}=100\) cycles/bit~\cite{liu2023dependent}.

% Parameter setting:~\cite{liu2023dependent}
% \begin{align*}
% f_c &\in \text{Uniform}[1.2\,\text{GHz}, 2.5\,\text{GHz}] \\
% h_c &= \text{Exp}(2.5 \cdot 10^{-7}) \\
% B &= 20\,\text{MHz}, \quad S = 0.5\,\text{MB} = 4 \cdot 10^6 \\
% \sigma^{2} &= -43\,\text{dBm} = 5.01 \cdot 10^{-8}\,\text{W}, \quad 
% C_{n} = 100\,\text{cycles/bit}, \quad 
% p_{c} = 0.2\,\text{W}
% \end{align*}

We consider 4 tasks in our experiment as illustrated in Figure \ref{fig:1}. The four tasks are distinct classification datasets: KMNIST (Task 1), MNIST (Task 2), FashionMNIST (Task 3), and QMNIST (Task 4), with thresholds $\tau_1=0.75$, $\tau_2=0.90$, $\tau_3=0.75$, and $\tau_4=0.85$. And we organized these tasks into a DAG with three layers: KMNIST in layer 1, MNIST and FashionMNIST in layer 2, and QMNIST in layer 3.

\subsection{Baselines}
We compare the total time and convergence performance of our proposed CoDa-FL with the following baseline approaches:
\begin{itemize}
  \item \textbf{EMD-Balanced Clustering} (EB): Clients are partitioned into equal-size groups based on EMD over their local data distributions.
  \item \textbf{K-Means Clustering} (KC): Clients are grouped by K-means on their label-distribution vectors to form training cohorts~\cite{cheng2021joint}.
  \item \textbf{Random Clustering} (RC): Clients are randomly assigned into fixed-size clusters to benchmark against structured grouping.
  % \item \textbf{Dynamic Greedy Selection} (DG/PS): In each round, a greedy heuristic picks the subset of clients whose combined update yields the largest immediate accuracy gain.
\end{itemize}

\subsection{Numerical Results}
% In our experiment, we consider four tasks, organized into a DAG with three layers: KMNIST is placed in layer 1, MNIST and FMNIST are grouped in layer 2, and QMNIST is assigned to layer 3.
All simulations results presented in this section have a 95\% confidence interval that is within 6\% of the observed mean.

\begin{figure}[htbp]
    \centering
    \captionsetup[subfigure]{font=footnotesize, skip=2pt}
    \begin{subfigure}[t]{0.49\linewidth}
        \centering
        \includegraphics[height=0.125\textheight,width=\linewidth]{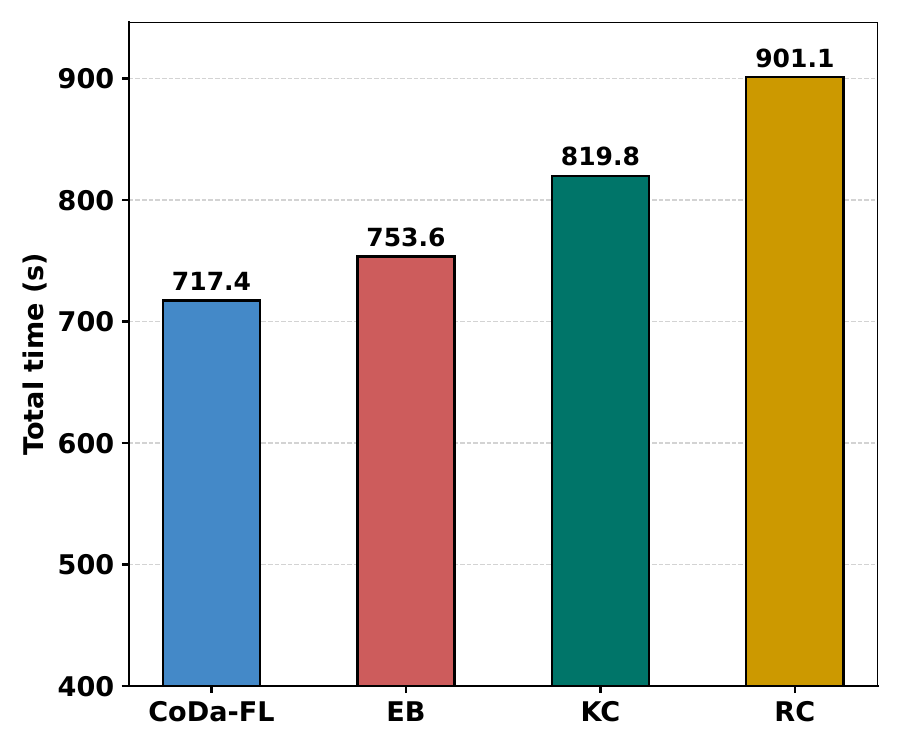}
        \caption{Total time to complete all tasks by different methods.}
         \label{fig:sub_time_models}
    \end{subfigure}
    \hfill
    \begin{subfigure}[t]{0.49\linewidth}
        \centering
        \includegraphics[height=0.125\textheight,width=\linewidth]{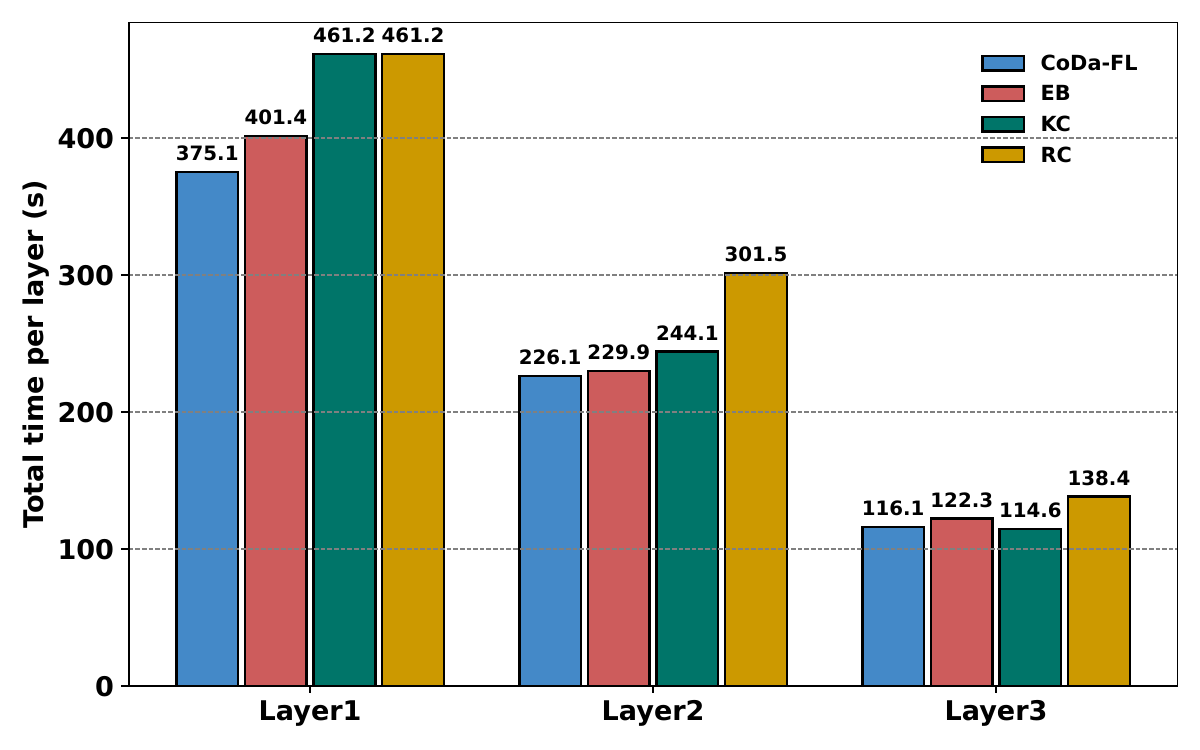}
        \caption{Total time for each layer in the DAG by different methods.}
        \label{fig:sub_time_layers}
    \end{subfigure}

    % \vspace{-8pt}
    \caption{Comparison of training time by different methods.}
    \label{fig:time_side_by_side}
\end{figure}

Figure~\ref{fig:sub_time_models} compares the total time required to complete all four tasks by each method. CoDa-FL achieves the shortest overall completion time among the cluster-based approaches. Specifically, CoDa-FL has an approximately 5.05\%
reduction in total time compared to EB, the closest baseline.
\vspace{0.1cm}

Note that, a non-cluster-based client selection mechanism that matches arbitrary combinations of clients to tasks achieves a total time of 427.5s to complete all learning tasks. This approach outperforms CoDa-FL in terms of total time efficiency. However, such mechanism is computationally prohibitive. For our case with $V=4$ and $U=100$, there are around $\frac{4^{100}}{24}$ combinations, while CoDa-FL’s EMD-based clustering and dependency-aware scheduling avoid such brute-force search, resulting in much lower algorithmic complexity without a severe sacrifice in performance.

Figure~\ref{fig:sub_time_layers} breaks down the training time by each layer of the task DAG. CoDa-FL exhibits consistent speedups over all baselines, with a significant advantage in layer 1 and layer 2. While KC performs better than CoDa-FL in layer 3, CoDa-FL's substantially shorter training times in the earlier layers result in the overall total time being the shortest among all models. %For the initial task CoDi completes in about 375.1s, which is ~6.5\% faster than EB and roughly 19\% faster than KC or RC. A similar trend is seen in Layer2 that the slowest of the two tasks finishes in under CoDi, slightly faster than under EB and dramatically faster than under the RC scheme. Even in the final layer, CoDi maintains an edge. 
These balanced improvements across all layers confirm that CoDa-FL’s efficiency gains are not confined to a particular task or phase; rather, the method accelerates each segment of overall FL training in a uniform way. This result also supports the robustness of CoDa-FL’s client grouping and scheduling strategy, across different training tasks. 

\begin{figure}[htbp]
    \centering
    \captionsetup[subfigure]{font=footnotesize, skip=2pt}

    % 第一行
    \begin{subfigure}{0.492\linewidth}
        \centering
        \includegraphics[height=0.15\textheight,width=\linewidth]{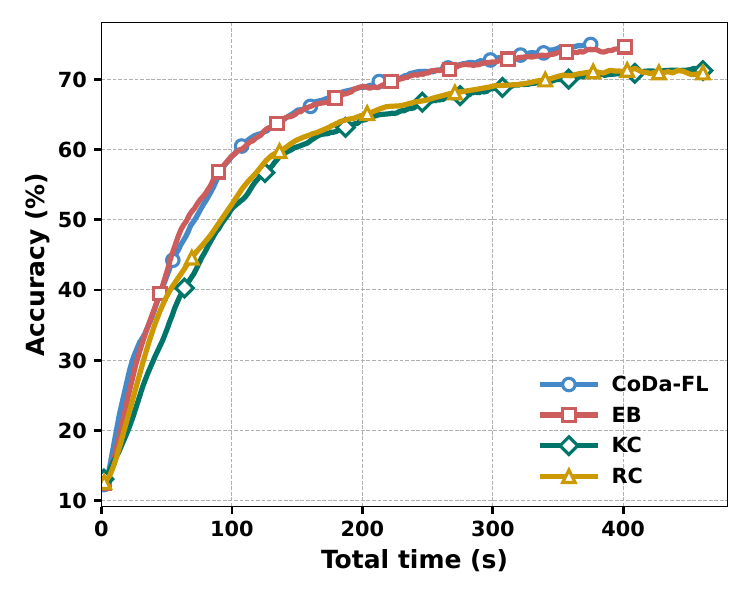}
        \caption{\textit{KMNIST}}
        \label{fig:kmnist}
    \end{subfigure}
    % \hfill
    \begin{subfigure}{0.492\linewidth}
        \centering
        \includegraphics[height=0.15\textheight,width=\linewidth]{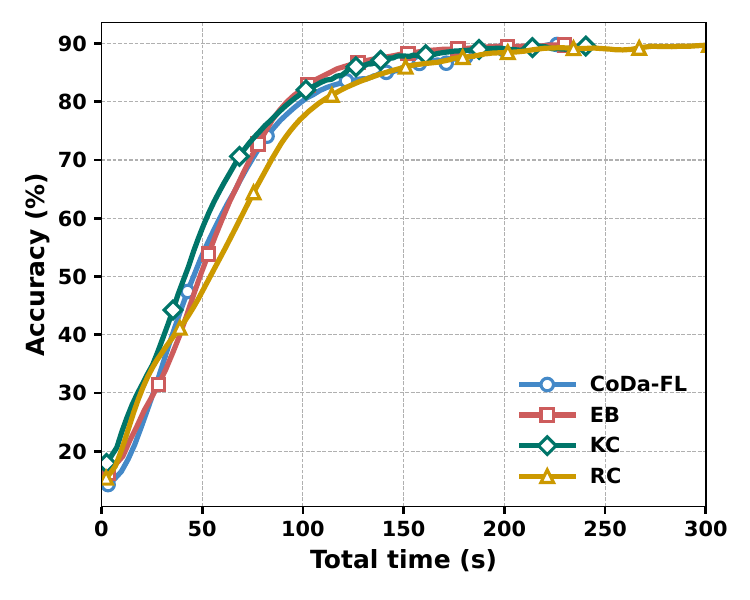}
        \caption{\textit{MNIST}}
        \label{fig:mnist}
    \end{subfigure}

    \vspace{0.8em} 

    % 第二行
    \begin{subfigure}{0.492\linewidth}
        \centering
        \includegraphics[height=0.15\textheight,width=\linewidth]{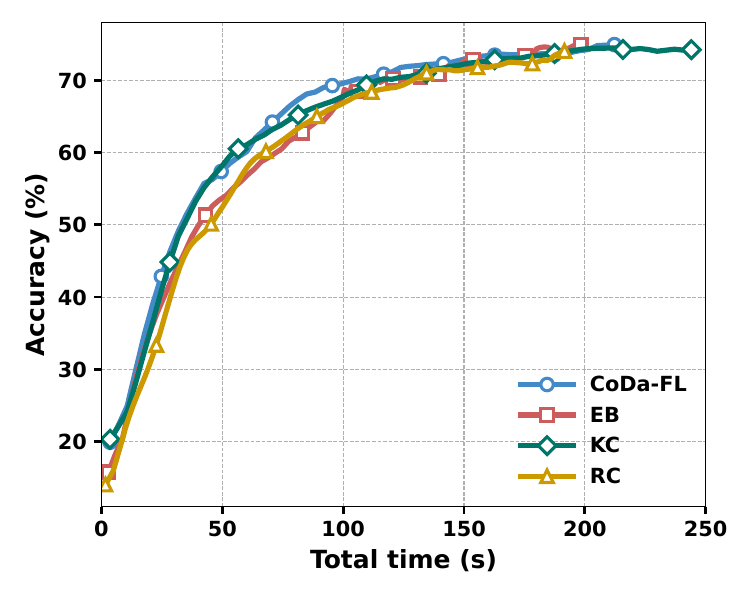}
        \caption{\textit{FashionMNIST}}
        \label{fig:fmnist}
    \end{subfigure}
    % \hfill
    \begin{subfigure}{0.492\linewidth}
        \centering
        \includegraphics[height=0.15\textheight,width=\linewidth]{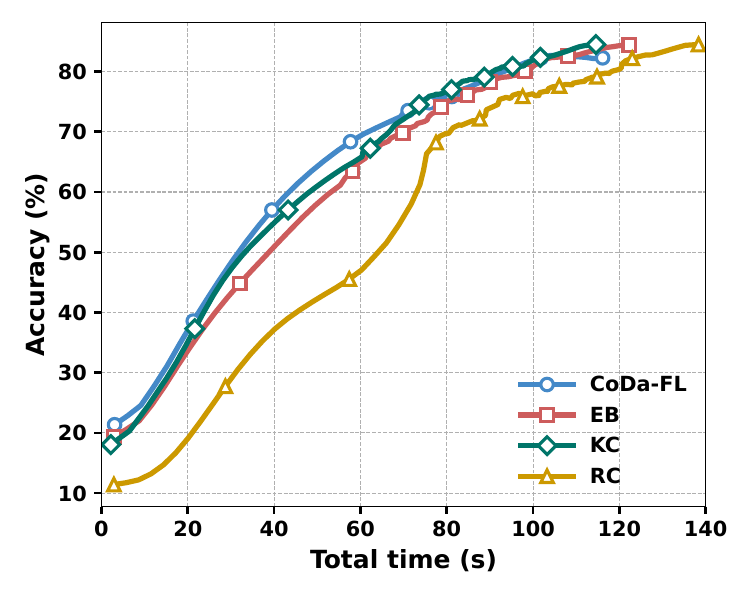}
        \caption{\textit{QMNIST}}
        \label{fig:qmnist}
    \end{subfigure}
    \caption{Accuracy vs. total time across four tasks.}
    \label{fig:acc_time_all}
\end{figure}

Figures~\ref{fig:acc_time_all} shows the accuracy achieved with respect to total time for each individual task. Across all four tasks, CoDa-FL reaches higher accuracy levels with less training time than other methods. For the FMNIST task in particular, CoDa-FL’s accuracy reaches 70\% within roughly 100s, whereas EB and KC require 120–150s to reach the same level. %Similar patterns hold for the parallel tasks MNIST and FashionMNIST: CoDi consistently achieves the target accuracy earlier than EB, KC, or RC, reflecting faster convergence. 

%Notably, the DG/PS method still converges slightly faster and reaches the accuracy target marginally ahead of CoDi on each task, as expected given its aggressive client-selection strategy. Nevertheless, the performance gap between CoDi and DG is relatively small, especially when contrasted with the much larger gap between CoDi and the other three baselines. In all cases, CoDi manages to combine rapid convergence with high final accuracy, nearly matching the sophisticated DG approach while distinctly outperforming the simpler clustering strategies.

 In summary, CoDa-FL reduces total completion time compared to traditional baselines (EB, KC, RC) while simultaneously avoiding the exponential computational complexity and runtime overhead associated with non-clustering selection methods. These results highlight CoDa-FL as the most practical and effective client selection strategy for dependent multi-task FL in MEC, delivering fast convergence and robust accuracy within a scalable and structured framework.
 
\ifCLASSOPTIONcaptionsoff
  \newpage
\fi

\section{Conclusion}
We proposed CoDa-FL, a Cluster-oriented and Dependency-aware framework to effectively address complex client selection challenges under the dependent multi-task assignment of FL in MEC environments. By identifying a key relationship between average EMD across clusters and the upper bound on required training rounds for each cluster-task pair, and by implementing a DAG-based task scheduling strategy, our approach significantly reduced the complexity of client selection and task assignment, thereby improving efficiency by optimizing the total time for all tasks. Through both theoretical analysis and numerical experiments, we demonstrated that our proposed approach significantly improves communication and computational efficiency over existing benchmarks.

% \section*{Acknowledgment}
% This work is partly supported by the Guangdong Provincial/Zhuhai Key Laboratory of Interdisciplinary Research and Application for Data Science, Project ...

\section*{Acknowledgment}
This work is partly supported by the Guangdong Provincial/Zhuhai Key Laboratory of 
Interdisciplinary Research and Application for Data Science, Project 2022B1212010006 and in part by Guangdong Higher Education Upgrading
Plan (2021-2025) UIC [R0400001-22] and [R0400024-22].
\bibliographystyle{IEEEtran}
\bibliography{ref}

\vfill


\end{document}